\documentclass[11pt]{article}

\usepackage{acl}
\usepackage{amsmath}
\usepackage{amssymb}
\usepackage{times}
\usepackage{latexsym}
\usepackage{xspace}
\usepackage{booktabs}
\usepackage{multirow}
\usepackage{graphicx}
\usepackage{subcaption}
\newcommand{\myparatights}[1]{\smallskip\noindent{\bf {#1}.}~}
\usepackage{float}
\usepackage{placeins}

\newcommand{\alg}{\text{STRIDE}\xspace}

\definecolor{cvprblue}{rgb}{0.21,0.49,0.74}
\definecolor{greyL}{RGB}{230,248,255}
\definecolor{yellow}{RGB}{255,246,234}
\definecolor{green}{RGB}{230,247,239}
\definecolor{blues}{RGB}{226,234,249}

\usepackage[most]{tcolorbox}
\usepackage{xcolor}

\newcommand{\stpos}[1]{\textcolor{red}{#1}}
\newcommand{\stweak}[1]{\textcolor{blue}{#1}}

\newtcolorbox{strideexample}[2][]{
  enhanced,
  breakable,
  colback=blue!2,
  colframe=blue!18,
  coltitle=blue!60!black,
  fonttitle=\bfseries\small,
  title={#2},
  boxrule=0.45pt,
  arc=1.5mm,
  left=1.5mm,
  right=1.5mm,
  top=1mm,
  bottom=1mm,
  before skip=5pt,
  after skip=5pt,
  #1
}

\usepackage[T1]{fontenc}

\usepackage[utf8]{inputenc}

\usepackage{microtype}

\usepackage{inconsolata}

\usepackage{graphicx}

\title{\alg: Strategic Trajectory Reasoning via Discriminative Estimation for Verifiable Reinforcement Learning}

\author{
\begin{tabular}{c}
Qinjian Zhao\textsuperscript{1,*},
Zhihao Dou\textsuperscript{2,*},
Zhang Dinggeng\textsuperscript{1,*},
Xiangyu Li\textsuperscript{3},
Chaoda Song\textsuperscript{2}, \\
Zhongwei Wan\textsuperscript{4},
Xinpeng Li\textsuperscript{2},
Yanyan Zhang\textsuperscript{2},
Kaijie Chen\textsuperscript{5},
Qingtao Pan\textsuperscript{2}, \\
Chengcheng Feng\textsuperscript{6},
Zhiqiang Gao\textsuperscript{1},
Xiaoyu Xia\textsuperscript{7}
\\[1ex]
\small
\textsuperscript{1}Kean University \quad
\textsuperscript{2}Case Western Reserve University \quad
\textsuperscript{3}University of Texas at Austin
\\
\small
\textsuperscript{4}The Ohio State University \quad
\textsuperscript{5}Tongji University \quad
\textsuperscript{6}Duke Kunshan University
\\
\small
\textsuperscript{7}Royal Melbourne Institute of Technology \quad
\textsuperscript{*}Equal contribution
\end{tabular}
}

\begin{document}
\maketitle
\begin{abstract}
Reinforcement Learning with Verifiable Rewards (RLVR) has become an effective post-training paradigm for improving the reasoning abilities of large language models. However, existing RLVR methods typically rely on final-answer correctness to assign trajectory-level rewards, providing sparse supervision and treating all tokens uniformly regardless of their actual contribution to reasoning. Although recent studies introduce intermediate signals such as process rewards, high-entropy tokens, and semantic uncertainty, these signals are often not inherently verifiable and may fail to distinguish beneficial strategic patterns from harmful ones. To address this limitation, we propose \alg (\textbf{S}trategic \textbf{T}rajectory \textbf{R}easoning with \textbf{D}iscriminative \textbf{E}stimation), a fine-grained RLVR framework that derives strategic reasoning supervision from verifiable outcomes. \alg contrasts successful and failed trajectories within each response group to estimate the outcome-discriminative preference of each $n$-gram strategic pattern, and further combines this signal with reasoning saliency entropy to identify decision-relevant strategic patterns. These patterns are assigned differentiated advantage values during RL optimization, enabling more precise credit assignment while preserving the verifiability of RLVR.
Extensive experiments demonstrate that \alg consistently improves reasoning performance across diverse models, tasks, and extended settings, including VLMs and agent-based systems.

\end{abstract}

\section{Introduction}
\label{sec:intro}

Reinforcement Learning with Verifiable Rewards (RLVR) \citep{shao2024deepseekmath,lambert2024tulu,yu2026dapo} has recently emerged as an effective post-training paradigm for enhancing the complex reasoning abilities of large language models (LLMs) across a wide range of domains \citep{zhang2025landscape,xiong2026ovd,dou2025plan,dong2025agentic}.
However, existing RLVR methods typically rely solely on the correctness of the final answer to assign rewards over the entire reasoning trajectory \cite{shao2024deepseekmath}. Such rule-based reward signals are inherently sparse, as they fail to provide fine-grained supervision for intermediate reasoning processes \citep{guo2025deepseek,cui2025process,chen2025discriminative}. Consequently, all tokens within a trajectory are often optimized uniformly during training regardless of their actual contribution to reasoning, which limits the model's ability to identify and reinforce decision-critical strategic patterns \citep{wang2025emergent,wang2026beyond}.

To address this issue, recent studies have introduced intermediate supervision signals from different perspectives during RL training \citep{wang2025emergent,yao2026prl,wang2026beyond,wang2024math,chen2025seed}. These signals include process reward models (PRMs) that evaluate intermediate reasoning steps \citep{yao2026prl,wang2024math}, unusually high-entropy tokens that indicate potential decision points \cite{wang2026beyond}, predefined strategic reasoning tokens \cite{wang2025emergent}, and response-level semantic uncertainty signals \cite{chen2025seed}. Based on these signals, different methods assign differentiated rewards or advantages to specific tokens, reasoning steps, or entire responses at different granularities, thereby providing more fine-grained optimization guidance beyond final-answer correctness.

\begin{figure*}[t]
    \centering
    \subfloat[Token-level entropy distribution]{\includegraphics[width=0.49\textwidth]{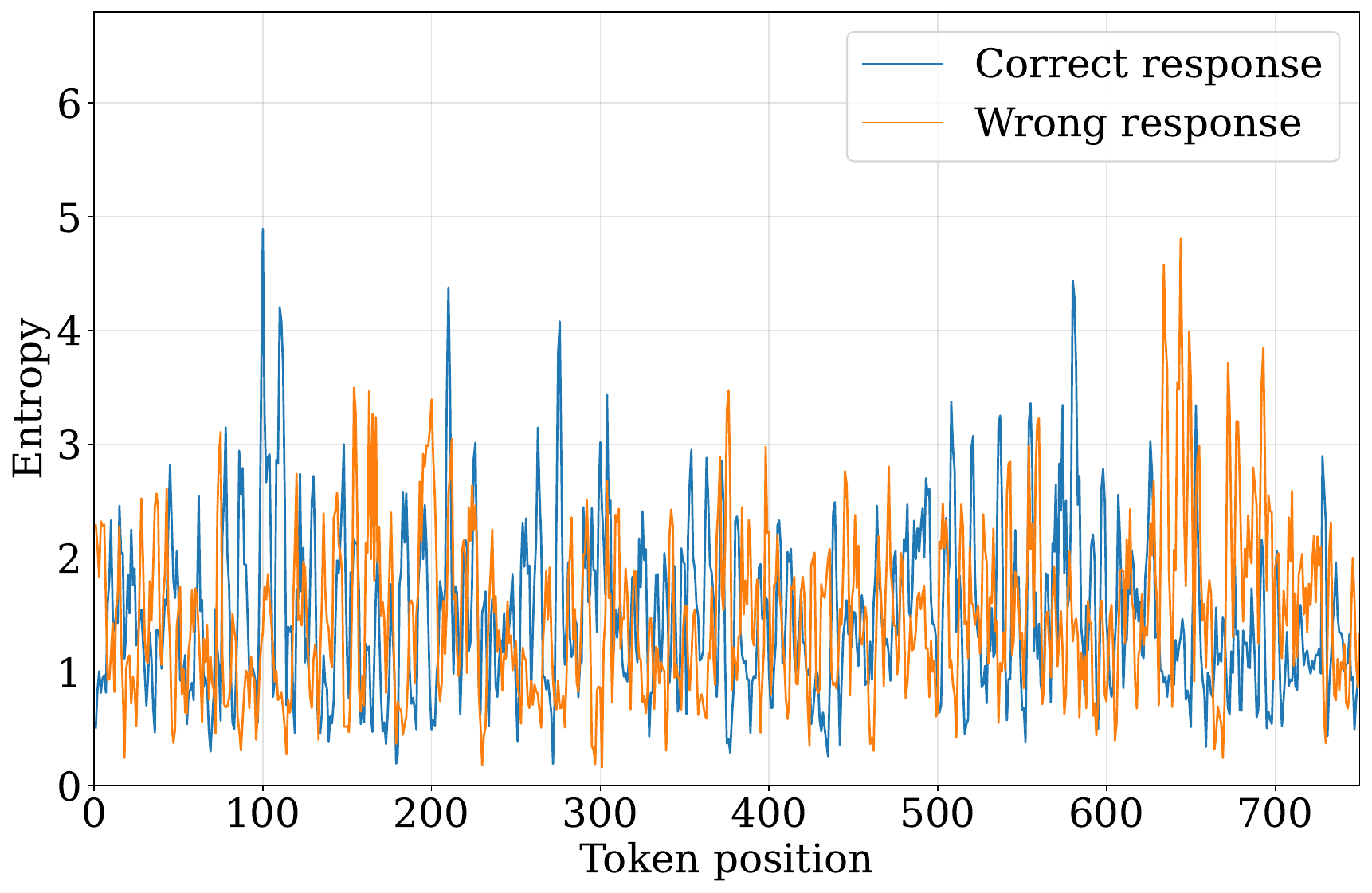}\label{fig:entr_c_w_line}}
    \hfill   
    \subfloat[Token-level entropy density distribution]{\includegraphics[width=0.49\textwidth]{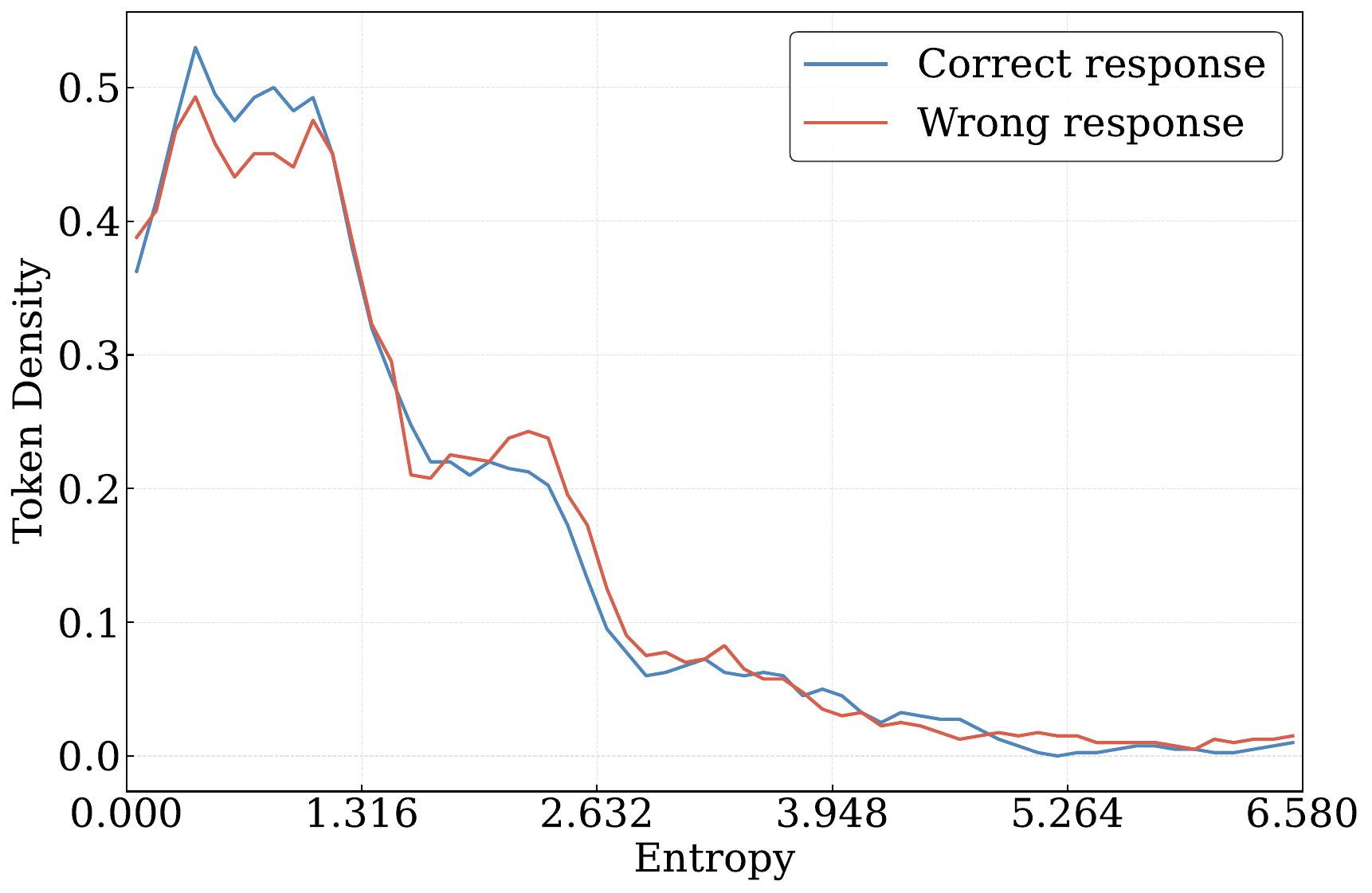}\label{fig:entr_c_w_d}}
    
    \caption{(a) Token-level entropy trajectories of correct and incorrect responses to the same questions show highly similar patterns, with nearly indistinguishable high-entropy peaks and frequencies. 
(b) The entropy density distributions of correct and incorrect responses also largely overlap, indicating that entropy alone cannot reliably determine whether a reasoning region contributes positively or negatively to the final outcome.}
    \label{fig:difference}
\end{figure*}

However, the intermediate signals used in these studies are often not inherently verifiable. Although they can identify uncertain or salient regions in reasoning trajectories, they cannot reliably determine whether these regions contribute positively or negatively to the final outcome.
As shown in Figure \ref{fig:entr_c_w_line}, for the same questions, correct and incorrect trajectories exhibit highly similar entropy patterns, with nearly indistinguishable high-entropy peaks and frequencies. Figure \ref{fig:entr_c_w_d} further confirms this through density distributions, showing that high- and low-entropy tokens follow similar frequency distributions in correct and incorrect trajectories. This suggests that entropy alone cannot distinguish beneficial reasoning regions from harmful ones.
Therefore, entropy-based signals can identify potentially important or strategic tokens \cite{wang2026beyond,pan2026prototype,qian2026demystifying}, but cannot determine their actual contribution to correctness. Similarly, even predefined strategic tokens \cite{wang2025emergent} are not necessarily beneficial.
We further verify this in Figure \ref{fig:entro_distri_token}. Tokens with high average entropy mostly correspond to reasoning-related words, but their distributions differ substantially between correct and incorrect trajectories, suggesting that such tokens may contribute either positively or negatively to the final answer.


Therefore, based on the above analysis, it is crucial to identify, in a verifiable manner, which strategic tokens truly contribute positively to the correct final answer. To bridge this gap, we propose \alg (\textbf{S}trategic \textbf{T}rajectory \textbf{R}easoning with \textbf{D}iscriminative \textbf{E}stimation), a fine-grained RLVR framework that derives supervision from verifiable outcomes. 
Specifically, \alg establishes a \textbf{D}iscriminative \textbf{S}trategic \textbf{R}easoning \textbf{R}einforcement \textbf{L}earning (\textbf{DSR-RL}) framework, which contrasts successful and failed reasoning trajectories within each response group to estimate the outcome-discriminative preference of each $n$-gram strategic pattern. Patterns that appear more frequently in successful trajectories are regarded as beneficial, whereas those more correlated with failed trajectories are treated as harmful. To avoid over-reinforcing routine continuations, \alg further integrates this discriminative signal with reasoning-saliency entropy, thereby identifying patterns that are both decision-relevant and verifiably aligned with final-answer correctness.
These strategic patterns are then assigned differentiated advantage values during RL optimization, enabling more precise credit assignment while preserving the verifiability of RLVR.
Extensive experimental results demonstrate that \alg achieves significant improvements across multiple model architectures and reasoning tasks. Moreover, \alg can be extended to VLM architectures and agent-related tasks, showing its broad applicability beyond text-only reasoning.

\section{Preliminaries and related work}

\subsection{Reinforcement Learning with Verifiable Rewards}

Reinforcement Learning with Verifiable Rewards (RLVR) improves LLM reasoning by optimizing models with automatically verifiable task rewards \cite{shao2024deepseekmath}.
However, outcome-level RLVR may encourage locally optimized responses without a coherent global plan \cite{wan2026srpo}, as it provides limited supervision over the intermediate reasoning process \cite{wang2025emergent,yao2026prl,wang2026beyond,wang2024math,chen2025seed}.
Existing methods primarily tackle this problem from two directions: designing alternative rollout strategies to derive more informative reward signals \cite{wan2026srpo,dou2025plan,chu2026redsearcher,he2025webseer,he2026iapo}, and supervising intermediate reasoning behaviors to provide denser guidance \cite{wang2025emergent,chen2025seed,wang2026beyond}.
Specifically, in the direction of constructing more informative rewards, PTA-GRPO \citep{dou2025plan} adopts plan-conditioned rollouts to build planning-aware rewards, while SRPO introduces a first-solution--reflection--refined-solution rollout and rewards concise yet effective self-correction \citep{wan2026srpo}. In the direction of intermediate-process supervision, HICRA \citep{wang2025emergent} amplifies learning signals on high-level planning tokens, whereas high-entropy-token methods restrict policy-gradient updates to critical forking tokens that steer reasoning directions \citep{wang2026beyond}.

\subsection{Preliminaries of Group Relative Policy Optimization (GRPO)}
\label{sec:grpo}

Group Relative Policy Optimization (GRPO) is a widely used RLVR algorithm for improving LLM reasoning. Compared with PPO \citep{schulman2017proximal}, GRPO removes the value model and estimates advantages from a group of sampled responses, making it more efficient for large-scale reasoning optimization.

Formally, for each question $q \in Q$, the old policy $\pi_{\theta_{\mathrm{old}}}$ samples $N$ responses $\{\mathfrak{o}_i\}_{i=1}^N$. Let $\mathcal{B}=\{q\sim Q,\{\mathfrak{o}_i\}_{i=1}^{N}\sim
\pi_{\theta_{\mathrm{old}}}(\cdot|q)\}$ and $L_i=|\mathfrak{o}_i|$.
For each token, define
\begin{equation}
\small
\begin{aligned}
\ell_{i,t}(\theta)
=
\min\Big(
\rho_{i,t} A_i,\,
\operatorname{clip}(\rho_{i,t},1-\epsilon,1+\epsilon) A_i
\Big),
\end{aligned}
\label{eq:grpo_surrogate}
\end{equation}
where
\begin{equation}
\small
\rho_{i,t}
=
\frac{
\pi_{\theta}(\mathfrak{o}_{i}^{t}\mid q,\mathfrak{o}_{i}^{<t})
}{
\pi_{\theta_{\mathrm{old}}}(\mathfrak{o}_{i}^{t}\mid q,\mathfrak{o}_{i}^{<t})
}.
\end{equation}
Then the GRPO objective is
\begin{equation}
\small
\begin{aligned}
J_{\mathrm{GRPO}}(\theta)
=
\mathbb{E}_{\mathcal{B}}
\Bigg[
\frac{1}{N}
\sum_{i=1}^{N}
\sum_{t=1}^{L_i}
\ell_{i,t}(\theta)
-
\beta D_{\mathrm{KL}}(\pi_{\theta}\|\pi_{\mathrm{ref}})
\Bigg].
\end{aligned}
\label{eq:grpo_objective}
\end{equation}
The response-level advantage is computed by normalizing rewards within the group:
\begin{equation}
\small
\begin{aligned}
A_i = \frac{r_i-\mu}{\sigma},\quad
\mu &= \frac{1}{N}\sum_{j=1}^{N} r_j, \quad
\sigma = \sqrt{\frac{1}{N}\sum_{j=1}^{N}(r_j-\mu)^2}.
\end{aligned}
\label{eq:grpo_advantage}
\end{equation}

In standard GRPO, the reward $r_i$ is usually determined by final-answer correctness. Although this outcome-level reward is verifiable and scalable, it provides sparse supervision for long CoT trajectories and does not explicitly evaluate intermediate reasoning quality. As a result, GRPO may still reward redundant or spurious reasoning as long as the final answer is correct.

\begin{figure}
    \centering
    \includegraphics[width=0.95\linewidth]{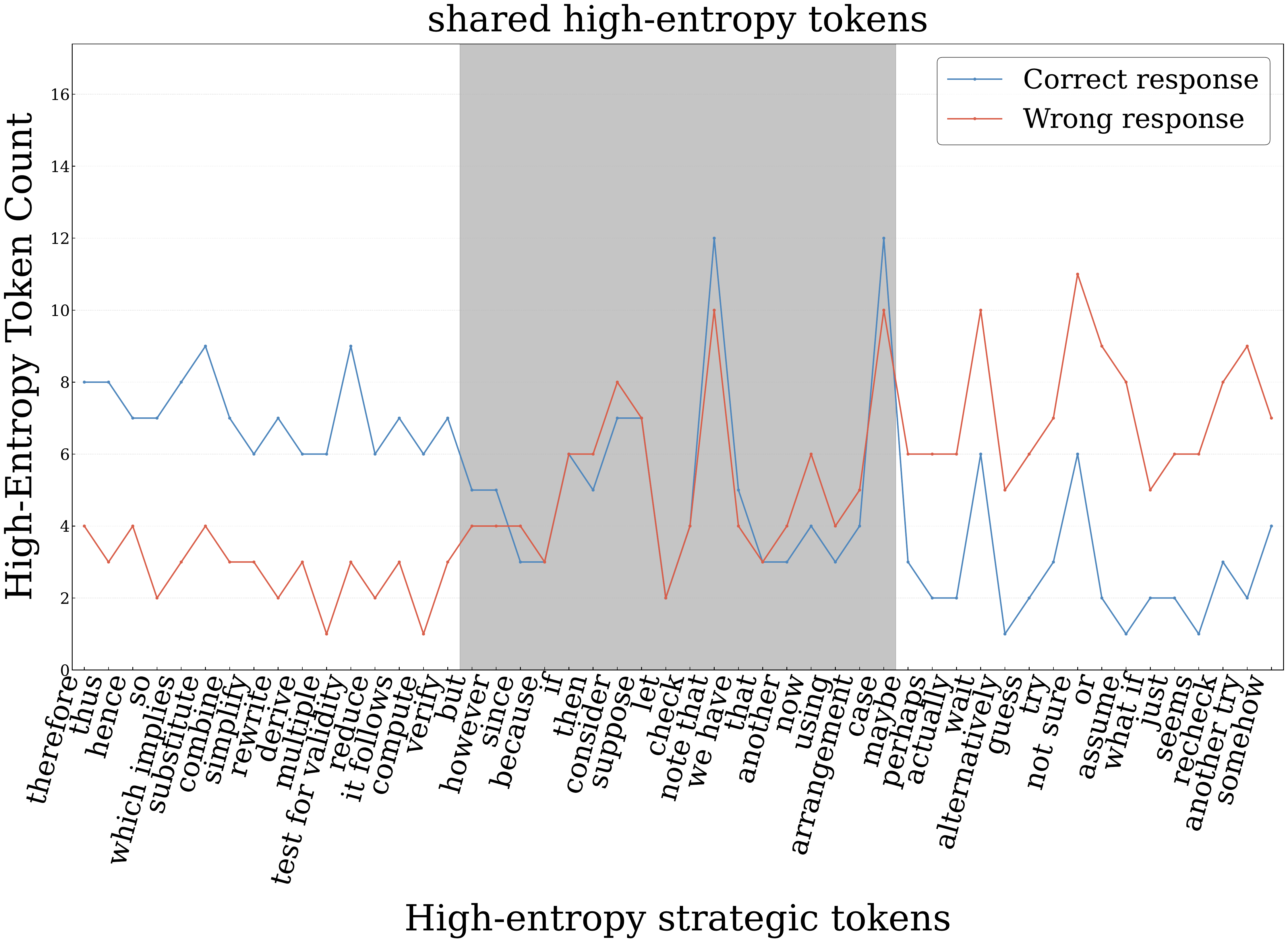}
    \caption{Distribution of high-entropy reasoning tokens in correct and wrong trajectories. The shaded regions highlight token ranges that are shared by both correct and wrong trajectories and exhibit similar frequencies.}
    \label{fig:entro_distri_token}
\end{figure}

\section{Approach of \alg}

\begin{figure*}
    \centering
    \includegraphics[width=0.9\linewidth]{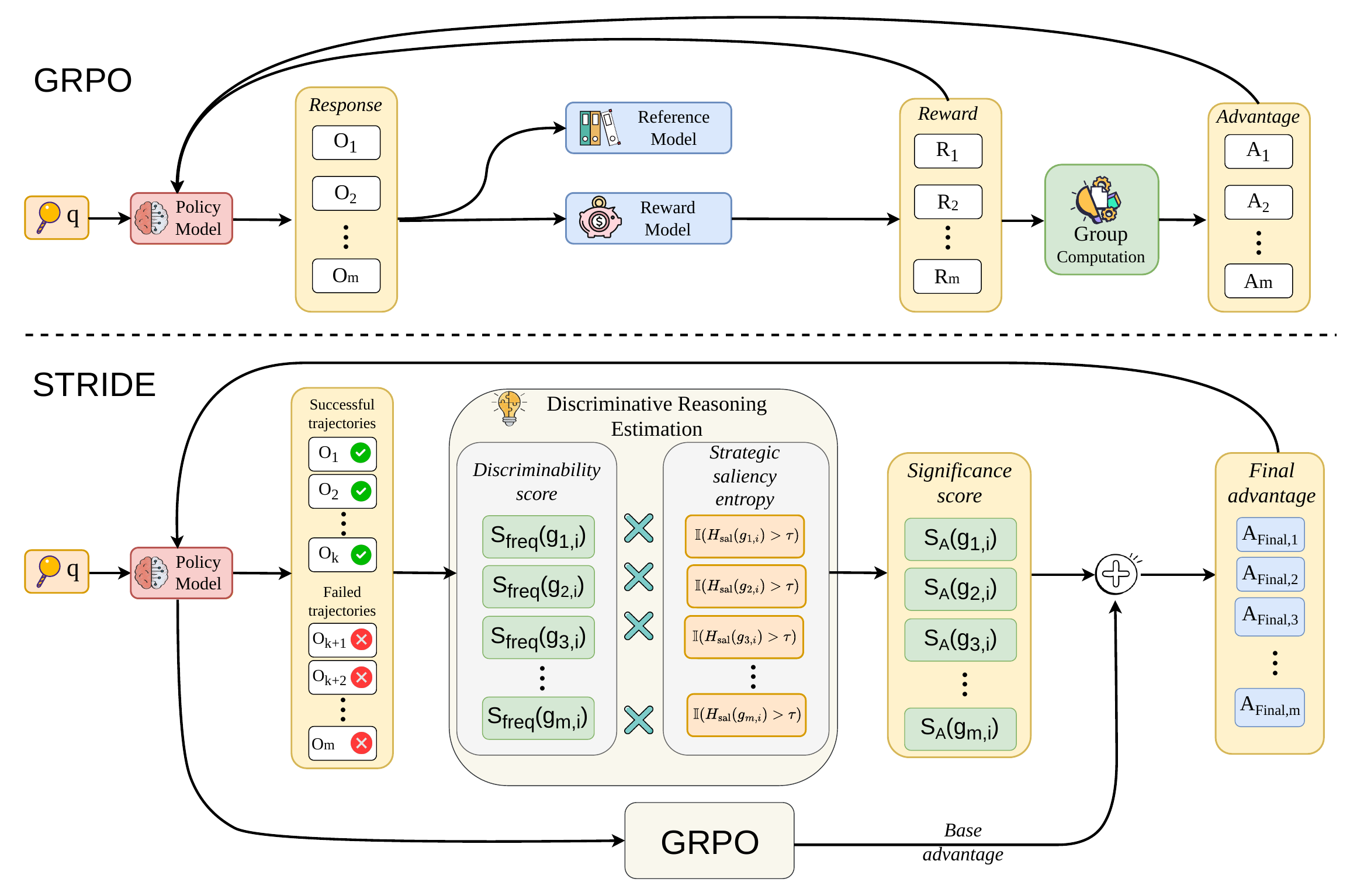}
    \caption{Overview of STRIDE. STRIDE contrasts successful and failed trajectories to identify outcome-discriminative strategic patterns, and assigns pattern-level advantages to reinforce beneficial patterns while penalizing harmful ones. More examples are provided in Appendix~\ref{sec:example}.}
    \label{fig:overall}
\end{figure*}

\alg aims to identify and optimize reasoning-critical regions within generated trajectories.
Unlike existing RLVR methods that uniformly optimize all tokens, \alg explicitly models the heterogeneous functional roles of tokens during reasoning and distinguishes whether specific strategic patterns contribute to successful or failed outcomes. To this end, \alg introduces the Discriminative Strategic Reasoning Reinforcement Learning (DSR-RL) framework in Sec.~\ref{sec:dsr}, with additional visualizations and analysis in Sec.~\ref{sec:empircal}. The overall workflow is shown in Fig.~\ref{fig:overall}.

\subsection{Discriminative Strategic Reasoning Reinforcement Learning (DSR-RL)}
\label{sec:dsr}

\subsubsection{Discriminative Reasoning Estimation}

Given a question $q$, the policy model $\pi_{\theta}$ samples a group of $m$ responses
$G=\{\mathfrak{o}_i\}_{i=1}^{m}$. Each response $\mathfrak{o}_i$ is generated as
$
\mathfrak{o}_i \sim \pi_{\theta}(\cdot \mid q),
$
where $\mathfrak{o}_i$ denotes the $i$-th sampled response.
For each response $\mathfrak{o}_i$, we represent it as a token sequence
$\mathfrak{o}_i = [\mathfrak{o}_{i,1}, \ldots, \mathfrak{o}_{i,L_i}]$, which contains both the reasoning trajectory and the final answer. Here, $\mathfrak{o}_{i,L_i}$ denotes the $L_i$-th token of response $\mathfrak{o}_i$.
Within each response group $G$, we partition the sampled responses into a successful subset $\mathcal{D}^{+}$ and a failed subset $\mathcal{D}^{-}$, such that
$G = \mathcal{D}^{+} \cup \mathcal{D}^{-}$.
These two subsets are defined as follows:
\begin{equation}
\mathcal{D}^{+}=\{\mathfrak{o}_i \mid \hat{y}_i = y\} \quad
\mathcal{D}^{-}=\{\mathfrak{o}_i \mid \hat{y}_i \neq y\},
\end{equation}
where $\hat{y}_i$ denotes the predicted answer of response $\mathfrak{o}_i$, and $y$ is the corresponding ground-truth answer.

To estimate whether an $n$-gram strategic pattern $g \in \mathfrak{o}_i$ is associated with correct or incorrect decision-making, we first define a frequency-based discriminability score $S_{\text{freq}}$ to quantify its outcome discriminability as follows:
\begin{equation}
S_{\text{freq}}(g)
=
\mathrm{clip}\left(
\log
\frac{
P(g|\mathcal{D}^{+})+\delta
}{
P(g|\mathcal{D}^{-})+\delta
},
-\epsilon, \epsilon
\right),
\end{equation}
where $P(g|\mathcal{D}^{+})$ and $P(g|\mathcal{D}^{-})$ denote the occurrence frequency of $g$ in successful and failed trajectories, respectively. 
$\delta$ is a smoothing constant for numerical stability, and the clipping threshold $\epsilon$ prevents rare $n$-gram patterns from producing excessively large log-ratio scores.
A large positive value of $S_{\text{freq}}(g)$ indicates that the corresponding strategic pattern is highly correlated with successful reasoning trajectories, whereas a smaller or negative value suggests that it is more associated with failed trajectories.



For a strategic pattern $g$, a high $S_{\text{freq}}(g)$ does not necessarily indicate that $g$ is causally helpful for deriving the correct answer. Although such patterns appear more frequently in the successful subset $\mathcal{D}^{+}$, they may still correspond to peripheral or non-critical \textit{execution patterns}, rather than decision-relevant \textit{strategic patterns}.
Therefore, it is still necessary to explicitly assess the actual importance of each strategic pattern $g$.

Inspired by previous studies~\cite{wang2026beyond,dong2025agentic}, high-entropy tokens in LLM reasoning are generally associated with more active, uncertain, and decision-relevant reasoning steps. Based on this empirical observation, we introduce \textbf{strategic saliency entropy} $H_{\text{sal}}(g)$, which quantifies the potential reasoning significance of an $n$-gram strategic pattern $g$, formulated as:

\begin{equation}
H_{\text{sal}}(g)
=
\frac{1}{n}
\sum_{t \in g}
H_t,
\end{equation}
where $n$ denotes the number of tokens in the strategic pattern $g$, and $H_t$ denotes the generation entropy at the token position $t$.
A higher value $H_{\text{sal}}(g)$ indicates that $g$ is more likely to contain decision-critical reasoning, while a lower value suggests continuation at the execution-level.

Based on the above analysis, we design a significance score $S_A(g)$ to quantify the contribution of a strategic pattern $g$ to the final outcome:

\begin{table*}[]
\centering
\caption{Performance comparison of different post-training methods across various base models. \textbf{Bold} indicates the best result within each block.}
\small
\setlength{\tabcolsep}{2pt} 
\begin{tabular}{c|c|cccccc|c}
\midrule[1.2pt]
Model                                  & Method                       & AIME24         & AIME25         & Math500        & AMC23          & Minerva        & Olympiad       & Avg.           \\ \hline
\multirow{6}{*}{Qwen2.5-7B-Instruct}   & Base                         & 12.24          & 3.52           & 62.40          & 52.75          & 39.16          & 34.52          & 34.10          \\
                                       & GRPO                         & 14.72          & 12.44          & 78.07          & 56.75          & 41.82          & 45.84          & 41.61          \\
                                       & DAPO                         & 15.64          & 13.07          & 78.25          & 55.31          & 42.21          & 46.31          & 41.80          \\
                                       & HICRA-GRPO                   & 16.73          & 13.19          & 77.58          & 55.41          & 42.42          & 46.74          & 42.01          \\
                                       & HET-DAPO                     & 16.96          & 13.77          & 78.55          & 56.30          & 42.97          & 44.53          & 42.18          \\
                                       & \alg                         & \textbf{18.46} & \textbf{15.59} & \textbf{80.17} & \textbf{59.69} & \textbf{45.92} & \textbf{48.45} & \textbf{44.71} \\ \hline
\multirow{6}{*}{Llama-3.1-8B-Instruct} & Base                         & 4.20           & 0.60           & 50.20          & 17.10          & 20.90          & 13.70          & 17.78          \\
                                       & GRPO                         & 9.25           & 0.73           & 54.22          & 25.63          & 26.40          & 21.32          & 22.92          \\
                                       & DAPO                         & 9.47           & 0.75           & 55.15          & 26.89          & 26.05          & 22.12          & 23.41          \\
                                       & HICRA-GRPO                   & 8.23           & 0.87           & 55.25          & 27.58          & 25.93          & 21.95          & 23.30          \\
                                       & HET-DAPO                     & 9.82           & 1.35           & 55.48          & 28.22          & 26.47          & 22.51          & 23.97          \\
                                       & \alg                         & \textbf{11.25} & \textbf{2.23}  & \textbf{57.19} & \textbf{31.43} & \textbf{28.82} & \textbf{24.29} & \textbf{25.87} \\ \hline
\multirow{6}{*}{Qwen3-4B-Base}         & Base                         & 9.40           & 5.30           & 63.80          & 38.90          & 28.30          & 30.70          & 29.40          \\
                                       & GRPO                         & 25.85          & 25.67          & 83.75          & 51.25          & 38.65          & 46.52          & 45.28          \\
                                       & DAPO                         & 27.60          & 25.34          & 86.48          & 52.35          & 40.67          & 47.32          & 46.63          \\
                                       & HICRA-GRPO                   & 26.07          & 26.32          & 86.43          & 54.94          & 40.23          & 48.55          & 47.09          \\
                                       & HET-DAPO                     & 26.65          & 26.12          & 86.96          & 53.74          & 40.69          & 49.48          & 47.27          \\
                                       & \alg                         & \textbf{28.24} & \textbf{28.76} & \textbf{87.18} & \textbf{56.52} & \textbf{43.28} & \textbf{51.45} & \textbf{49.24} \\ \midrule[1.2pt]
\end{tabular}
\label{tab:various_method}
\end{table*}

\begin{table*}[]
\caption{
Ablation analysis on \alg, where Qwen2.5-7B-Instruct is considered as base model. \textbf{Bold} is best per block.
}
\centering
\small
\setlength{\tabcolsep}{5.0pt}
\begin{tabular}{l|cccccc|c}
\midrule[1.2pt]
Variant 
& AIME24 
& AIME25 
& Math500 
& AMC23 
& Minerva 
& Olympiad 
& Avg. \\
\midrule

\alg w/o $S_{\mathrm{freq}}$
& 15.24 & 14.56 & \textbf{85.65} & 54.27 & 40.19 & 43.23 
& 42.19 \\

\alg w/o Entropy Filter
& \textbf{18.52} & 11.07 & 76.35 & 54.39 & 40.55 & 46.74 
& 41.27 \\

\alg w/o N-gram Modeling
& 13.73 & 11.52 & 76.96 & 54.67 & 40.82 & 44.63 
& 40.39 \\

\alg
& 18.46
& \textbf{15.59} 
& 80.17
& \textbf{59.69} 
& \textbf{45.92} 
& \textbf{48.45} 
& \textbf{44.71} \\

\midrule[1.2pt]
\end{tabular}
\label{tab:qwen25_ablation}
\end{table*}

\begin{equation}
S_A(g) = \mathbb{I}\!\left(H_{\text{sal}}(g)\ge \tau\right) \cdot S_{\text{freq}}(g),
\end{equation}
where $\mathbb{I}(\cdot)$ is an indicator function that equals $1$ if the condition holds and $0$ otherwise, and $\tau$ is a threshold hyperparameter. 
This design ensures that a strategic pattern $g$ is considered a meaningful strategic pattern only when its saliency entropy $H_{\text{sal}}(g)$ is sufficiently high. 
Once activated by the entropy condition, the value of $S_A(g)$ is determined by $S_{\text{freq}}(g)$: a larger positive $S_{\text{freq}}(g)$ yields a larger $S_A(g)$, indicating that $g$ is more beneficial to correct reasoning and should receive stronger reward. 
Conversely, a smaller or negative $S_{\text{freq}}(g)$ leads to a negative $S_A(g)$, suggesting that $g$ is more associated with failed reasoning and should be penalized. 
When $S_A(g)$ is close to or equal to zero, the pattern is regarded in practice as having limited influence on the final outcome. More empirical analysis is provided in Sec.~\ref{sec:empircal}.

\subsubsection{Strategic Advantage Reweighting and Model Optimization}

After obtaining the significance score $S_A(g)$ for each strategic pattern $g \in G$, we use it to compute the corresponding advantage value during the RL optimization process. Following conventional RLVR in Eq.\ref{eq:grpo_advantage}, we first compute a base advantage
$A_{\text{base}}$ from the response group $G$, where the advantage is determined
by the correctness of the final answer.
Building upon this, we further combine it with the significance score $S_A(g)$ to generate final advantage $A_{\text{final}}$ as follows:

\begin{equation}
    A_{\text{final}} = A_{\text{base}} + S_A(g).
\end{equation}

Here, $S_A(g)$ serves as an $n$-gram-level advantage term that is only applied to activated strategic patterns. 
By adding $S_A(g)$ to the base advantage, \alg further reweights the uniform outcome-level reward in conventional RLVR, assigning additional reward to strategic patterns that contribute to correct reasoning and stronger penalty to those associated with failed reasoning. Based on the final advantage $A_{\text{final}}$, we optimize the policy model following the RLVR objective in Eq.~\ref{eq:grpo_objective}.

\subsection{Visualization and Statistical Analysis}
\label{sec:empircal}

We first analyze high-entropy reasoning tokens in correct and wrong trajectories. As shown in Figure~\ref{fig:entro_distri_token}, many high-entropy tokens correspond to reasoning-related expressions, including transition, verification, and uncertainty words. However, shared high-entropy tokens appear with similar frequencies in both correct and wrong trajectories. This suggests that entropy can locate potentially important reasoning regions, but cannot determine whether they are beneficial or harmful, which may lead to inaccurate credit assignment.

We further visualize high-frequency reasoning tokens in Figure~\ref{fig:difference}. Correct trajectories contain more derivation- and verification-related tokens, while wrong trajectories include more uncertainty-oriented tokens. Nevertheless, many frequent tokens are shared by both types of trajectories (see Appendix \ref{sec:shared_token_all}), indicating that frequency alone is not sufficiently outcome-discriminative. These observations motivate STRIDE to combine frequency-based discriminative estimation with reasoning-saliency entropy, enabling more precise identification of decision-relevant and outcome-aligned strategic patterns.

\begin{figure*}[t]
    \centering
    \subfloat[High-frequency tokens in correct trajectories]{\includegraphics[width=0.47\textwidth]{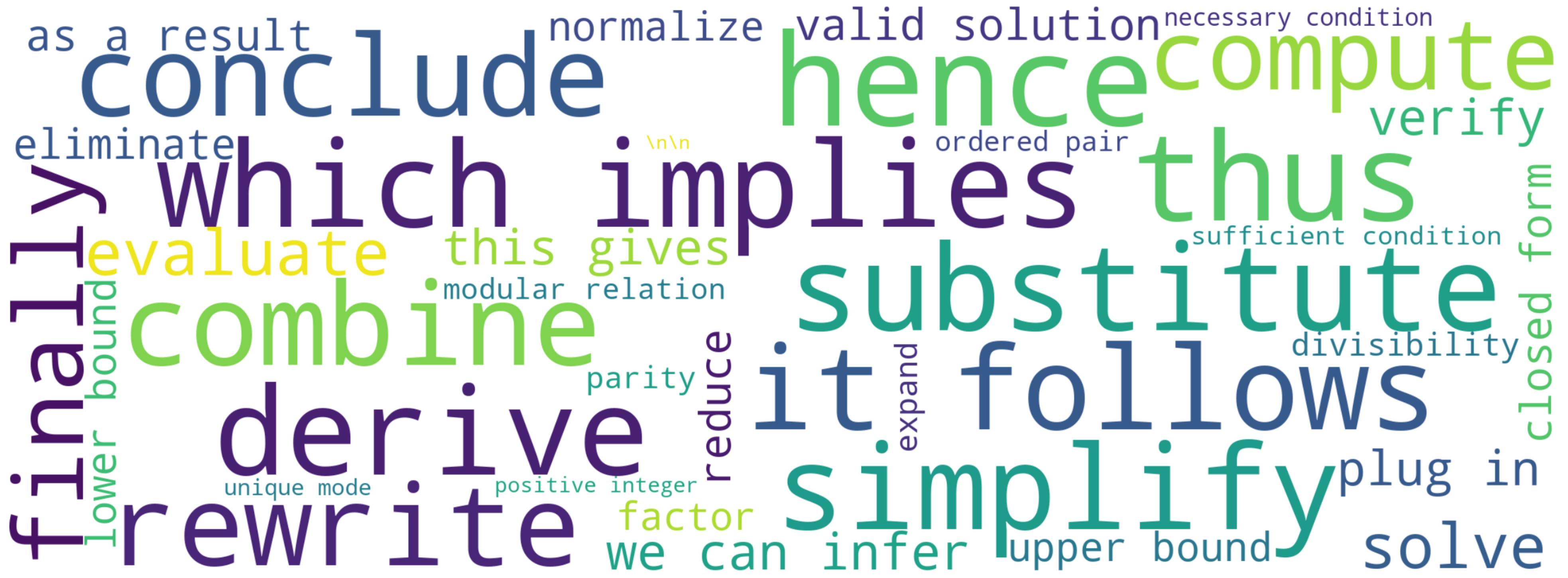}\label{fig:cot-example}}
    \hfill   
    \subfloat[High-frequency tokens in wrong trajectories]{\includegraphics[width=0.47\textwidth]{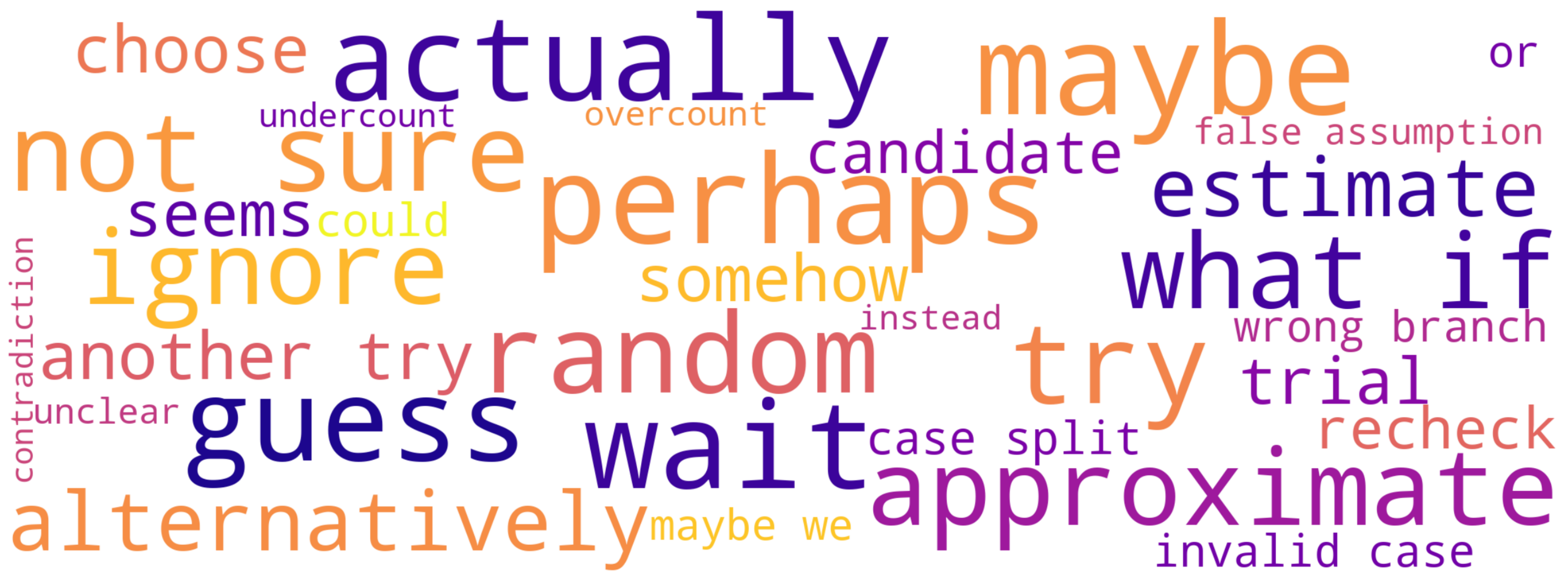}\label{fig:ours-example}}
    \hfill   
    
    \caption{Visualization of high-frequency reasoning tokens in different trajectories.
(a) High-frequency tokens in correct trajectories.
(b) High-frequency tokens in wrong trajectories.}
    \label{fig:difference}
\end{figure*}

\section{Experiment}

\subsection{Experiment setting}

\myparatights{Base Models}
To evaluate \alg, we employ three base models from different model families and scales, including Llama-3.1-8B-Instruct \citep{dubey2024llama}, Qwen2.5-7B-Instruct \citep{bai2025qwen2}, and Qwen3-4B-Base \citep{yang2025qwen3}. This setup enables a comprehensive assessment of the robustness and generalizability of \alg across diverse architectures. The training details are provided in Appendix~\ref{sec:details}. For the vision-language setting, we further extend our experiments to Qwen2.5-VL-7B \citep{bai2025qwen2}, which serves as the base VLM.

\myparatights{Training Datasets and Benchmarks}
For RL training, we sample 30K problems from DeepMath~\citep{he2025deepmath}, which provides graded difficulty annotations and is rigorously decontaminated to avoid benchmark leakage. To further examine the generalization of \alg across different task domains, we also conduct training on multimodal tasks~\citep{wan2026srpo} and agent benchmarks~\citep{xi2024agentgymevolvinglargelanguage}.
We evaluate \alg on mathematical reasoning benchmarks, including AIME24, AIME25, MATH500, Minerva, Olympiad, and AMC23, and report the average accuracy over 64 independent runs. In addition, we assess its general-purpose multimodal reasoning ability on MMMU-Pro~\citep{yue2025mmmu}, MMMU~\citep{yue2024mmmu}, and EMMA~\citep{standley2023extensible}, as well as its scientific reasoning performance on MMK-12~\citep{meng2025mm}. Finally, to evaluate agentic decision-making and interactive reasoning, we test \alg on representative agent benchmarks, including WebShop~\citep{yao2023webshopscalablerealworldweb}, ALFWorld~\citep{shridhar2021alfworldaligningtextembodied}, and BabyAI~\citep{chevalierboisvert2019babyaiplatformstudysample}.

\myparatights{Baseline}
We compare \alg with the base model and several RLVR baselines, including GRPO~\citep{shao2024deepseekmath}, DAPO \cite{yu2026dapo}, HICRA-GRPO \cite{wang2025emergent}, and HET-DAPO \cite{wang2026beyond}. To ensure a fair comparison, all methods are trained on the same SFT and RL datasets, with differences only in their method-specific SFT enhancement components. In addition, we keep the number of sampled responses identical across all RLVR methods, resulting in comparable training costs.

Due to space limitations, we provide additional experimental results, including test-time scaling, training dynamics, and statistical significance analysis in Appendix~\ref{sec:more_exp}.

\begin{table}[]
\centering
\caption{Comparison between \alg and other approaches on General-Benchmark and Science Benchmark, using Qwen2.5-VL-7B as the base model.}
\setlength{\tabcolsep}{1.5pt} 
\footnotesize
\begin{tabular}{l|cccccc}
\midrule[1.2pt]
\textbf{Method}     & \textbf{MMMU-Pro} & \textbf{MMMU} & \textbf{EMMA} & \textbf{Phys} & \textbf{Chem} & \textbf{Bio}  \\ \hline
Base                & 30.5              & 54.1          & 20.2          & 45.4          & 56.4          & 54.0          \\
MM-Eureka           & 37.6              & 55.2          & 23.5          &56.2          & 65.2          & 65.2         \\
SRPO                & 42.3              & 57.1          & 29.6          & 60.6          & 70.4          & 69.5        \\
\alg & \textbf{44.2}     & \textbf{58.7} & \textbf{31.5} & \textbf{61.9} & \textbf{71.5} & \textbf{72.3} \\ \midrule[1.2pt]
\end{tabular}
\label{tab:mllm}
\end{table}

\begin{table*}[]
\centering
\small
\caption{Comparison between \alg and other approaches on agent tasks, using Qwen2.5-VL-7B as the base model.}
\begin{tabular}{l|l|ccc|c}
\midrule[1.2pt]
Method                       & Type           & WebShop       & ALFWorld      & BabyAI        & Avg.          \\ \hline
LLM-Planner~\citep{song2023llmplannerfewshotgroundedplanning}                  & Prompting      & 18.9          & 68.5          & 82.5          & 56.6          \\
AgentTraj-L-SFT~\citep{xi2024agentgymevolvinglargelanguage}                   & SFT            & 73.5          & 83.2          & 74.2          & 77.0          \\
RWR~\citep{10.1145/1273496.1273590}                                             & Offline RL     & 68.2          & 76.5          & 82.1          & 75.6          \\
DPO~\citep{rafailov2024directpreferenceoptimizationlanguage}                    & Offline RL     & 75.3          & 86.5          & 78.3          & 80.0          \\
PPO~\citep{schulman2017proximalpolicyoptimizationalgorithms}                    & Online RL      & 68.5          & 83.5          & 69.8          & 73.9          \\
AgentSTaR~\citep{xi2024agentgymevolvinglargelanguage}                           & Offline RL     & 76.5          & 88.6          & 82.7          & 82.6          \\ \hline
HET-DAPO~\citep{wang2026beyond}                                                 & Token-level RL & 77.1          & 88.4          & 83.0          & 82.8          \\
\alg                                                                            & Token-level RL & \textbf{78.8} & \textbf{89.6} & \textbf{84.5} & \textbf{84.3} \\ \midrule[1.2pt]
\end{tabular}
\label{tab:agentic}
\end{table*}

\subsection{Performance of \alg}

Table~\ref{tab:various_method} compares \alg with representative post-training methods across different base models and mathematical reasoning benchmarks. Overall, \alg consistently achieves the best performance across all models and benchmarks, demonstrating the robustness and general applicability of our method. On Qwen2.5-7B-Instruct, \alg improves the average score from 42.18 of the strongest baseline HET-DAPO to 44.71, yielding a gain of 2.53 points. Similar improvements are observed on Llama-3.1-8B-Instruct, where \alg outperforms HET-DAPO by 1.90 points on average, and on Qwen3-4B-Base, where \alg further improves the average score from 47.27 to 49.24. 

This suggests that \alg can improve both standard mathematical reasoning and harder problem-solving scenarios. These results demonstrate that the proposed discriminative strategic reasoning optimization is effective across different model families and model scales, and can provide stable improvements over both trajectory-level RL methods and token-level RL baselines.

\subsection{Ablation analysis}

\begin{figure*}[t]
    \centering
    \begin{subfigure}{0.32\textwidth}
        \centering
        \includegraphics[width=0.95\linewidth]{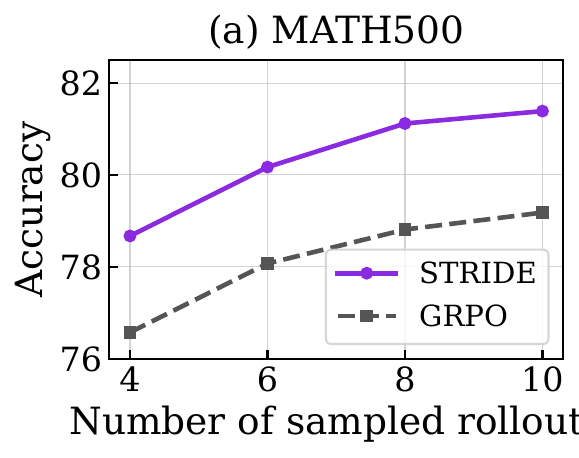}
        \caption{MATH500}
    \end{subfigure}
    \hfill
    \begin{subfigure}{0.32\textwidth}
        \centering
        \includegraphics[width=0.95\linewidth]{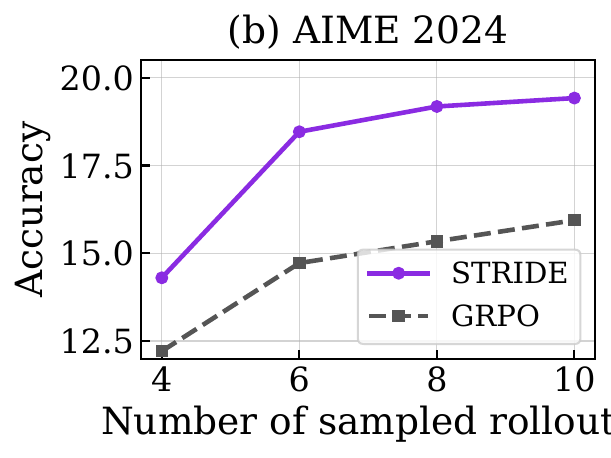}
        \caption{AIME 2024}
    \end{subfigure}
    \hfill
    \begin{subfigure}{0.32\textwidth}
        \centering
        \includegraphics[width=0.95\linewidth]{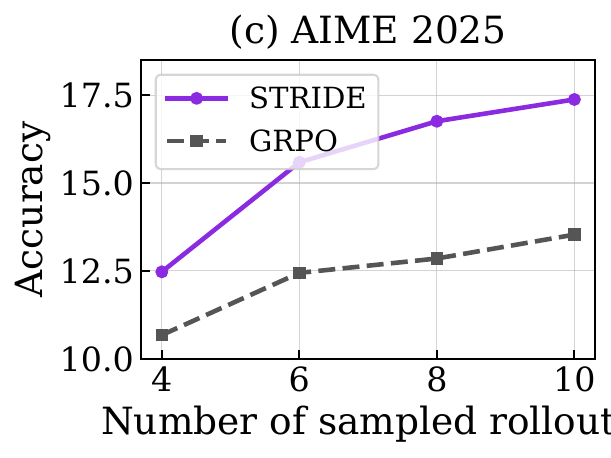}
        \caption{AIME 2025}
    \end{subfigure}

    \begin{subfigure}{0.32\textwidth}
        \centering
        \includegraphics[width=0.95\linewidth]{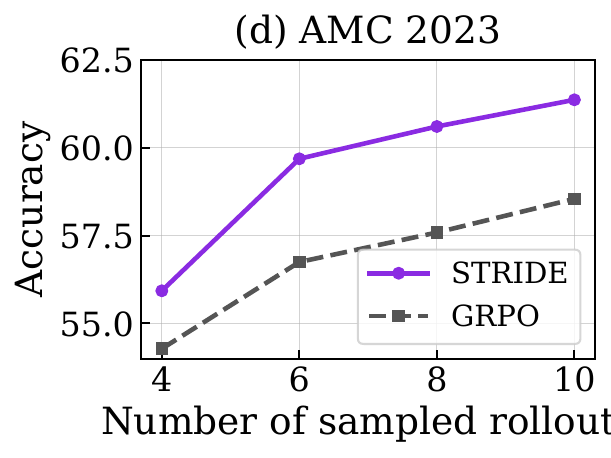}
        \caption{AMC 2023}
    \end{subfigure}
    \hfill
    \begin{subfigure}{0.32\textwidth}
        \centering
        \includegraphics[width=0.95\linewidth]{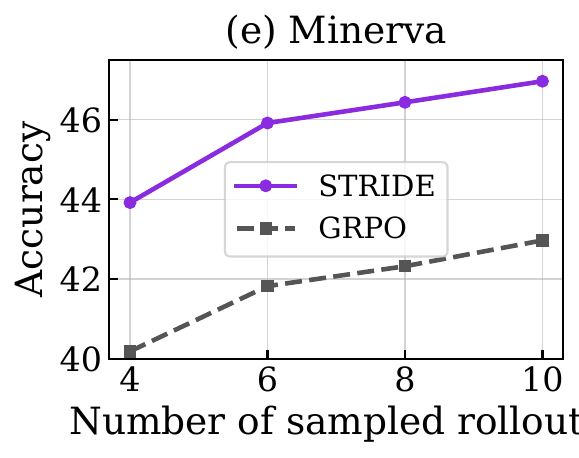}
        \caption{Minerva}
    \end{subfigure}
    \hfill
    \begin{subfigure}{0.32\textwidth}
        \centering
        \includegraphics[width=0.95\linewidth]{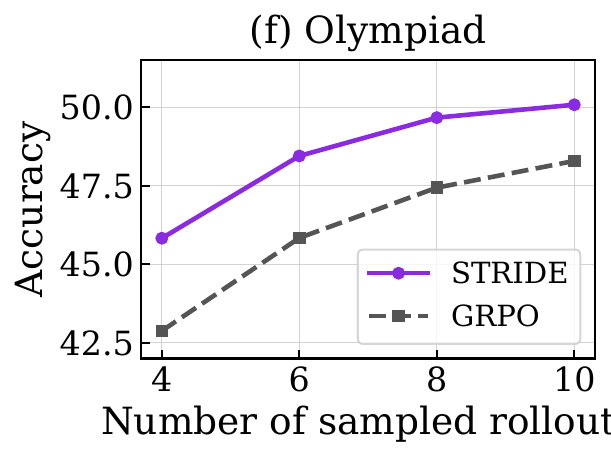}
        \caption{Olympiad}
    \end{subfigure}

    \caption{Sensitivity analysis of \alg with respect to the number of sampled rollouts, where Qwen2.5-7B-Instruct is used as the base model.}
    \label{fig:senstive_analysis}
\end{figure*}

Table~\ref{tab:qwen25_ablation} presents the ablation analysis of \alg on Qwen2.5-7B-Instruct. Removing any key component leads to a clear performance drop, confirming the effectiveness of each design. Without $S_{\mathrm{freq}}$, the average score decreases from 44.71 to 42.46, showing that frequency-based discriminative estimation is important for distinguishing beneficial and harmful reasoning patterns. Removing the entropy filter also reduces the average score to 42.31, indicating that reasoning-saliency information helps identify decision-relevant tokens. The largest drop occurs when removing $n$-gram modeling with single token, where the average score falls to 41.61. This suggests that modeling strategic reasoning patterns at the phrase level is crucial for fine-grained credit assignment. Overall, the full \alg achieves the best average performance, clearly demonstrating that these components are highly complementary.

\subsection{Generalization of \alg and baselines on multimodal reasoning tasks.}
Beyond text-only settings, we further evaluate \alg on multimodal, general-domain, and scientific benchmarks, including MMMU-Pro \citep{yue2025mmmu}, MMMU \citep{yue2024mmmu}, EMMA \citep{standley2023extensible}, and MMK-12 \citep{meng2025mm}, as shown in Table \ref{tab:mllm}. We adopt SRPO \citep{wan2026srpo} and MM-Eureka \citep{meng2025mm} as baselines, and follow SRPO’s SFT/RL data setup for cold-start initialization and subsequent RL training.
As shown in Table \ref{tab:mllm}, \alg consistently outperforms all baselines on both general-domain and scientific multimodal benchmarks. Compared with the base Qwen2.5-VL-7B model, \alg achieves substantial gains across all tasks, improving MMMU-Pro from 30.5 to 44.2, EMMA from 20.2 to 31.5, and Bio from 54.0 to 72.3. Moreover, \alg also surpasses strong RL-based baselines, including MM-Eureka and SRPO. Compared with SRPO, \alg obtains further improvements of 1.9 on MMMU-Pro, 1.6 on MMMU, 1.9 on EMMA, and 2.8 on Bio, demonstrating its effectiveness in enhancing both general multimodal reasoning and scientific reasoning capabilities.

\subsection{Generalization of \alg on agent tasks}
To further examine the generalizability of \alg beyond standard reasoning benchmarks, we extend our evaluation to interactive decision-making scenarios using agent tasks from AgentGym~\citep{xi2024agentgymevolvinglargelanguage}. Specifically, we use 7,160 AgentGym trajectories for training, where each trajectory contains a task instruction, environment observations, model actions, and the final task outcome. We evaluate \alg on three representative agent benchmarks: WebShop~\citep{yao2023webshopscalablerealworldweb}, ALFWorld~\citep{shridhar2021alfworldaligningtextembodied}, and BabyAI~\citep{chevalierboisvert2019babyaiplatformstudysample}. To ensure a fair comparison, all RL-based methods are trained on the same trajectory data.

Table~\ref{tab:agentic} reports the performance of \alg on interactive decision-making tasks from AgentGym. Overall, \alg achieves the best average score of 84.3 across WebShop, ALFWorld, and BabyAI. Compared with the strongest offline RL baseline AgentSTaR, \alg improves the average score from 82.6 to 84.3, with consistent gains on all three tasks. It also outperforms the token-level RL baseline HET-DAPO by +1.5 on average, showing that its discriminative token-level optimization generalizes to interactive agent scenarios.

\subsection{Sensitivity analysis}

Figure~\ref{fig:senstive_analysis} reports the sensitivity analysis of STRIDE with respect to the number of sampled rollouts. Across all six mathematical reasoning benchmarks, STRIDE consistently outperforms GRPO under different rollout budgets, demonstrating its robustness to the sampling scale. As the number of rollouts increases from 4 to 10, both methods generally achieve better accuracy, indicating that more sampled responses provide richer exploration signals for RL optimization. However, STRIDE maintains a clear advantage over GRPO at every rollout setting, especially on challenging benchmarks such as AIME 2024, AIME 2025, and Olympiad. This suggests that STRIDE can more effectively identify and reinforce beneficial strategic patterns from sampled trajectories. Moreover, the performance gain gradually saturates when the rollout number increases from 8 to 10, showing that STRIDE can already achieve strong performance with a moderate sampling budget.

\section{Conclusion}

In this paper, we propose \alg, a fine-grained RLVR framework that improves reasoning by identifying outcome-discriminative strategic patterns from verifiable successful and failed trajectories. By combining frequency-based estimation with reasoning-saliency entropy, STRIDE assigns differentiated advantages to beneficial and harmful strategic patterns, enabling more precise credit assignment than standard outcome-level RLVR. Extensive experiments across mathematical reasoning, multimodal reasoning, and agent tasks show that STRIDE consistently outperforms strong baselines, demonstrating its effectiveness, robustness, and broad generalizability.

\section*{Limitations}

Although \alg demonstrates consistent improvements across mathematical reasoning, multimodal reasoning, and agent tasks, several directions remain worth further exploration. First, our current study mainly focuses on tasks with automatically verifiable outcomes, which provide reliable signals for discriminative reasoning estimation. Extending this framework to more open-ended generation scenarios, where automatic verification is less straightforward, is an interesting future direction. Second, \alg estimates strategic patterns from sampled response groups. While our experiments show that the method remains effective under moderate rollout budgets, further improving the efficiency of pattern estimation could make the framework more broadly applicable. Finally, this work adopts $n$-gram patterns as a simple and effective unit for fine-grained credit assignment. Future work may explore more flexible semantic units to further improve interpretability and generalization.

\section*{Ethics statement}

This work focuses on improving the reasoning ability of large language models through verifiable reinforcement learning. 
STRIDE derives fine-grained supervision from automatically verifiable outcomes and assigns differentiated optimization signals to strategic patterns that are more associated with successful or failed trajectories. 
While this framework is designed to improve reasoning reliability and credit assignment, the resulting models may still inherit biases, factual inaccuracies, or unsafe behaviors from their pretrained backbones and training data. 
Moreover, stronger reasoning capabilities may be misused in inappropriate or harmful applications if deployed without proper safeguards.

To reduce these risks, our experiments are conducted on public and commonly used reasoning, multimodal, and agent benchmarks, with a primary focus on mathematical reasoning, scientific reasoning, and structured decision-making tasks rather than sensitive personal or harmful content. 
The proposed method relies on verifiable task outcomes instead of subjective human preference labels, which helps reduce annotation bias and improve reproducibility. 
Nevertheless, STRIDE is not intended to replace human oversight in high-stakes domains. 
We encourage future deployments to include careful safety evaluation, bias analysis, and domain-specific human review before applying the method to real-world systems involving sensitive decisions.


\bibliography{custom}

\clearpage
\appendix




\section{Appendix}
\label{sec:appendix}

\subsection{Experimental parameter setup}
\label{sec:details}

All RL experiments were conducted with GRPO as the base RLVR algorithm. For each input question, the policy model generated a group of 6 responses during rollout. STRIDE then contrasted successful and failed trajectories within each group to estimate outcome-discriminative $n$-gram strategic patterns. We set the rollout batch size to 640 and used a global batch size of 160 for policy optimization, with a micro-batch size of 2 per GPU. During training-time generation, we used temperature $=1.0$ and top-$p=1.0$; during validation, we used temperature $=0.6$ and top-$p=0.95$. The optimizer was AdamW with a learning rate of $1.0 \times 10^{-6}$, weight decay of $1.0 \times 10^{-2}$, and a warmup ratio of 0. The KL coefficient was set to $1.0 \times 10^{-2}$. For STRIDE, the $n$-gram length ranged from 2 to 5, the weighting coefficient was 0.1, and the entropy threshold was 2.5. We set both the maximum prompt length and maximum response length to 2048 tokens. Unless otherwise stated, all hyperparameters were kept the same across models.


\subsection{Extra experiment}
\label{sec:more_exp}

\subsubsection{Results of test-time scaling}

\begin{figure*}[h]
    \centering
    \includegraphics[width=0.95\textwidth]{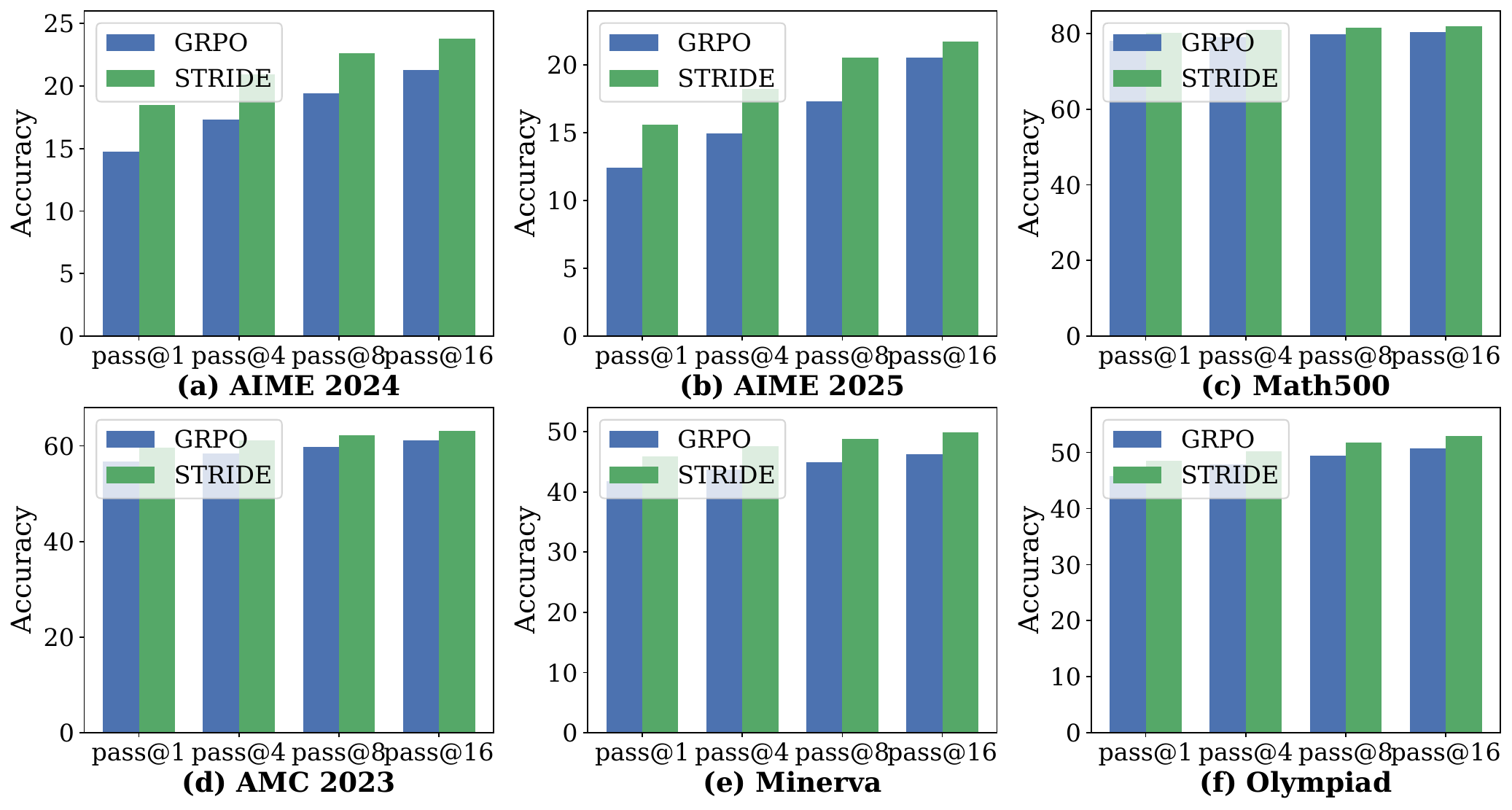}
    \caption{
    Effect of scaling test-time compute on six mathematical reasoning benchmarks.
    We report Pass@K results of GRPO and \alg with Qwen2.5-7B-Instruct as the base model.
    }
    \label{fig:passk_all_benchmarks}
\end{figure*}

Figure~\ref{fig:passk_all_benchmarks} analyzes the effect of scaling test-time compute by increasing the number of sampled rollouts. 
Across all six mathematical reasoning benchmarks, \alg consistently outperforms GRPO under different Pass@$K$ settings, demonstrating that our method maintains stable advantages as the sampling budget increases. 
The gains are particularly clear on challenging benchmarks such as AIME 2024, AIME 2025, and Olympiad, where \alg achieves higher accuracy than GRPO across all evaluated $K$ values. 
This indicates that \alg improves not only single-sample reasoning performance, but also the overall quality of sampled reasoning trajectories. 
Overall, these results show that \alg scales effectively with test-time compute and better leverages additional rollout budgets than GRPO.

\subsubsection{Training Dynamics of \alg}

\begin{figure*}[h]
    \centering

    \begin{subfigure}{0.32\textwidth}
        \centering
        \includegraphics[width=\linewidth]{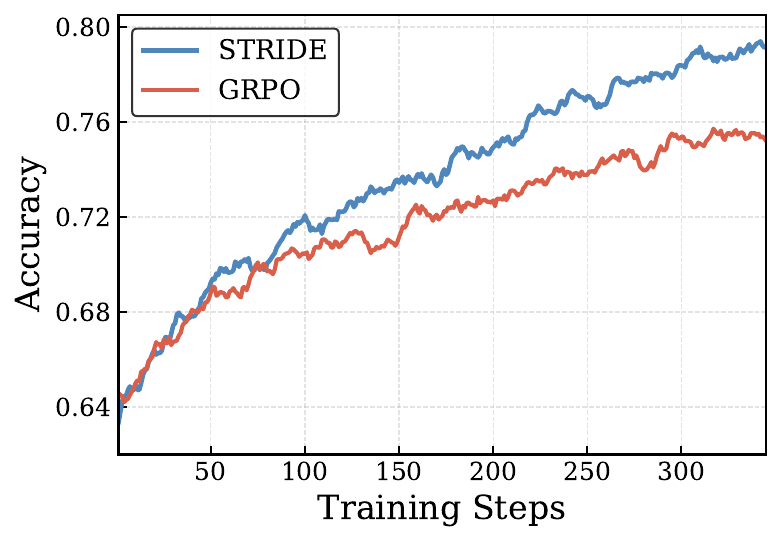}
        \caption{Accuracy reward}
    \end{subfigure}
    \hfill
    \begin{subfigure}{0.32\textwidth}
        \centering
        \includegraphics[width=\linewidth]{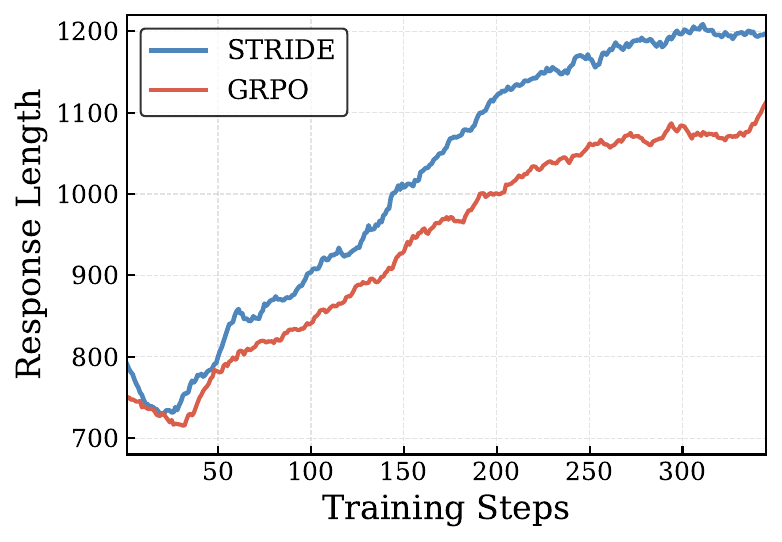}
        \caption{Response length}
    \end{subfigure}
        \hfill
    \begin{subfigure}{0.32\textwidth}
        \centering
        \includegraphics[width=\linewidth]{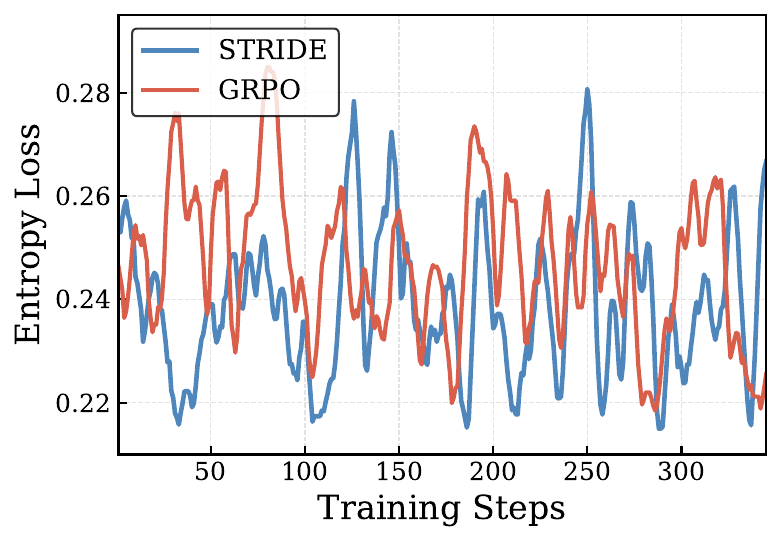}
        \caption{Entropy loss}
    \end{subfigure}

    \caption{
    Training Dynamics of \alg with Qwen2.5-7B-Instruct.
    }
    \label{fig:traing_rehensive_analysis}
\end{figure*}

Figure~\ref{fig:traing_rehensive_analysis} illustrates the training dynamics of \alg compared with GRPO on Qwen2.5-7B-Instruct. 
As training progresses, \alg achieves consistently higher accuracy rewards than GRPO, indicating more effective policy optimization. 
Meanwhile, \alg produces longer responses, suggesting that the model learns to generate more detailed reasoning trajectories during training. 
In terms of entropy loss, both methods exhibit fluctuating trends, while \alg remains within a comparable range to GRPO, showing that the performance improvement does not come from unstable exploration behavior. 
Overall, these results demonstrate that \alg improves reasoning performance steadily while maintaining stable training dynamics.

\subsubsection{Statistical Significance Analysis}

\begin{table}[h]
\centering
\caption{Pairwise significance analysis between \alg and GRPO across different models and mathematical reasoning benchmarks. $\Delta$ denotes the performance improvement of \alg over GRPO in percentage points.}
\small
\setlength{\tabcolsep}{1pt}
\begin{tabular}{l|l|l|c|c}
\toprule
Model & Baseline & Task & $\Delta$ (pt) & $p$ / sig \\
\midrule
\multirow{6}{*}{Qwen2.5-7B-Instruct}
& GRPO & MATH500  & +2.10 & $0.004^{**}$ \\
& GRPO & AIME24   & +3.74 & $0.003^{**}$ \\
& GRPO & AIME25   & +3.15 & $0.005^{**}$ \\
& GRPO & AMC23    & +2.94 & $0.021^{*}$ \\
& GRPO & Minerva  & +4.10 & $<0.001^{***}$ \\
& GRPO & Olympiad & +2.61 & $0.007^{**}$ \\
\midrule

\multirow{6}{*}{Llama-3.1-8B-Instruct}
& GRPO & MATH500  & +2.97 & $0.004^{**}$ \\
& GRPO & AIME24   & +2.00 & $0.006^{**}$ \\
& GRPO & AIME25   & +1.50 & $0.018^{*}$ \\
& GRPO & AMC23    & +5.80 & $<0.001^{***}$ \\
& GRPO & Minerva  & +2.42 & $0.032^{*}$ \\
& GRPO & Olympiad & +2.97 & $0.005^{**}$ \\
\midrule

\multirow{6}{*}{Qwen3-4B-Base}
& GRPO & MATH500  & +3.43 & $<0.001^{***}$ \\
& GRPO & AIME24   & +2.39 & $0.006^{**}$ \\
& GRPO & AIME25   & +3.09 & $0.004^{**}$ \\
& GRPO & AMC23    & +5.27 & $<0.001^{***}$ \\
& GRPO & Minerva  & +4.63 & $<0.001^{***}$ \\
& GRPO & Olympiad & +4.93 & $<0.001^{***}$ \\

\bottomrule
\end{tabular}
\label{tab:pairwise_significance}
\end{table}
Table~\ref{tab:pairwise_significance} reports the pairwise significance analysis between \alg and GRPO. Across all three base models and six mathematical reasoning benchmarks, \alg consistently achieves positive improvements over GRPO, indicating stable gains rather than isolated improvements. Most results reach strong statistical significance, with many cases showing $p<0.001$. For Qwen2.5-7B-Instruct and Llama-3.1-8B-Instruct, \alg also delivers consistent and statistically significant improvements, further demonstrating the robustness of the proposed optimization across different model families.

\subsection{Visualization of Shared Strategic Tokens}
\label{sec:shared_token_all}

\begin{figure*}[t]
    \centering
    \includegraphics[width=0.65\linewidth]{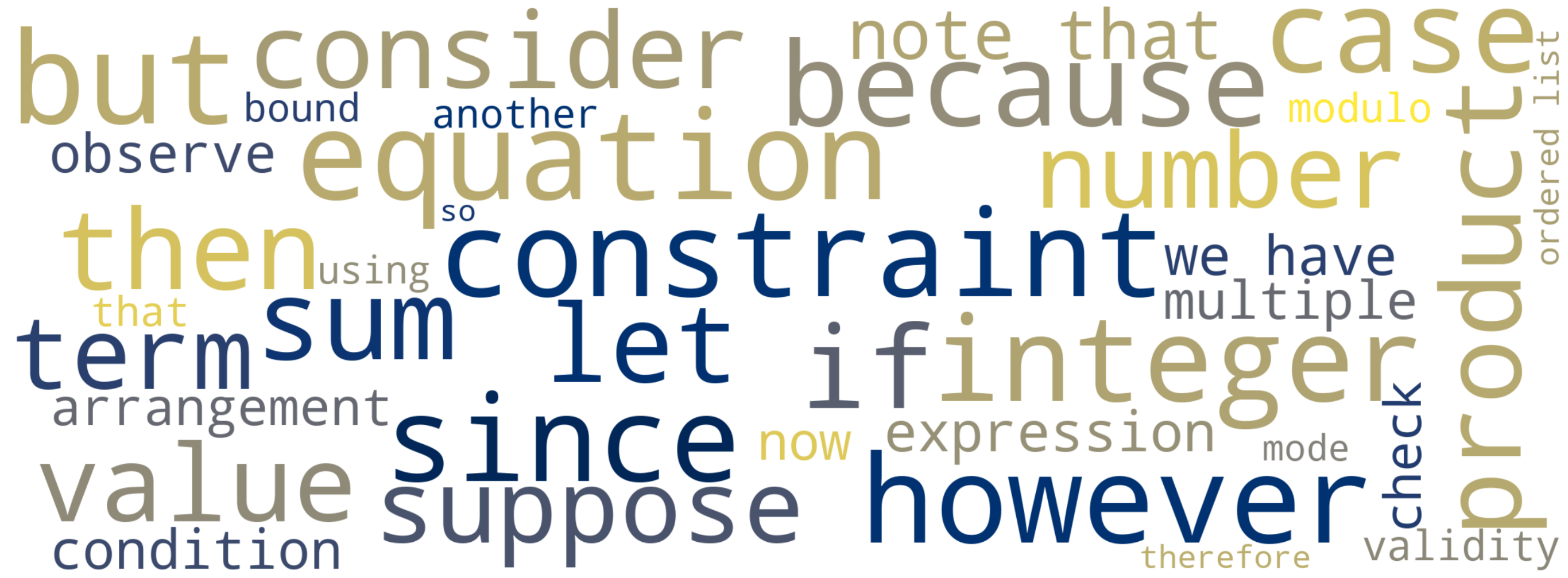}
    \caption{High-frequency tokens shared by correct and wrong trajectories, indicating that frequent reasoning tokens are not necessarily outcome-discriminative.}
    \label{fig:shared_token}
\end{figure*}

The shared token set can be seen in Fig \ref{fig:shared_token}.

\FloatBarrier


\clearpage

\subsection{Examples of Strengthened and Weakened Patterns}
\label{sec:example}
This section presents examples of strengthened and weakened 5-gram strategic patterns. The shown response is a rollout generated during the RL training process of \alg. In the response, red-highlighted phrases denote strategic reasoning patterns with high positive significance scores, which are reinforced by \alg, while blue-highlighted phrases denote strategic reasoning patterns with low or negative significance scores, which are weakened during reinforcement learning.

\begin{strideexample}{Example 1: Strengthened and weakened 5-gram strategic patterns}
\small

\textbf{Question.}
Every morning Aya goes for a $9$-kilometer-long walk and stops at a coffee shop afterwards.
When she walks at a constant speed of $s$ kilometers per hour, the walk takes her 4 hours,
including $t$ minutes spent in the coffee shop. When she walks $s+2$ kilometers per hour,
the walk takes her 2 hours and 24 minutes, including $t$ minutes spent in the coffee shop.
Suppose Aya walks at $s+\frac{1}{2}$ kilometers per hour. Find the number of minutes the walk
takes her, including the $t$ minutes spent in the coffee shop.

\vspace{2pt}
\textbf{Response.}
Let the stopping time be $t$ minutes. A tempting but incorrect approach is to
\stweak{guess decimal time from notation} and treat $2$ hours $24$ minutes as $2.24$ hours.
Instead, we first \stpos{derive equations from shared condition}: the total time equals walking
time plus the same coffee-shop time.

Since the equations use hours, we \stpos{convert minutes into hour units}, so the stopping time is
$\frac{t}{60}$ hours. Thus, for the first speed,
\[
\frac{9}{s}+\frac{t}{60}=4.
\]
For the second speed, we should not \stweak{approximately convert twenty four minutes}; rather,
$2$ hours $24$ minutes equals $\frac{12}{5}$ hours. Hence,
\[
\frac{9}{s+2}+\frac{t}{60}=\frac{12}{5}.
\]

Now the key reasoning step is to \stpos{compare equations under same variable}. The same $t$
appears in both equations, so we should not \stweak{ignore shared coffee shop time} or
\stweak{solve two cases independently first}. Therefore, we
\stpos{subtract equations to eliminate time}:
\[
\frac{9}{s}-\frac{9}{s+2}
=
4-\frac{12}{5}
=
\frac{8}{5}.
\]

At this point, another wrong path would be to \stweak{try random interpolation between speeds}
or \stweak{assume speed proportional to total}. However, since the distance is fixed, we
\stpos{use reciprocal speed time relation}. This gives
\[
9\left(\frac{1}{s}-\frac{1}{s+2}\right)=\frac{8}{5},
\]
which simplifies to
\[
\frac{18}{s(s+2)}=\frac{8}{5}.
\]
Then we \stpos{simplify equation and solve positive} speed:
\[
s=\frac{5}{2}.
\]

Next, we \stpos{substitute value back to compute} the fixed stopping time:
\[
\frac{9}{5/2}+\frac{t}{60}=4.
\]
Since $\frac{9}{5/2}=3.6$, it follows that $t=24$ minutes. Here we should not
\stweak{forget fixed time after solving}, because the question asks for the total time including
the coffee shop.

For the new speed,
\[
s+\frac{1}{2}=3.
\]
The walking time is $9/3=3$ hours, or $180$ minutes. Finally, we
\stpos{verify total includes stopping time} and \stpos{conclude final minutes including stop}:
\[
180+24=204.
\]

\textbf{Answer.}
\[
\boxed{204}
\]

\end{strideexample}

\begin{strideexample}{Example 2: Strengthened and weakened 5-gram strategic patterns}
\small

\textbf{Question.}
A list of positive integers has the following properties:
the sum of the items in the list is $30$; the unique mode of the list is $9$; and the median
of the list is a positive integer that does not appear in the list itself. Find the sum of the
squares of all the items in the list.

\vspace{2pt}
\textbf{Response.}
Sort the list in nondecreasing order. Since the median is required to be an integer that does
not appear in the list, we first \stpos{check parity of list length}. If the list had odd length,
then the median would be the middle item, which must appear in the list. This is impossible.
Therefore, we \stpos{conclude list length is even}.

Since the unique mode is $9$, the number $9$ must occur at least twice. A tempting but
incorrect approach is to \stweak{guess many small repeated numbers}; however, any other
number repeated as often as $9$ would destroy the unique mode condition. Hence we
\stpos{consider cases by nine count}.

First suppose that $9$ appears three times. Then these three terms already contribute
$27$, leaving only $3$ for all remaining positive integers. If there is one remaining term, the
list would have the form
\[
3,9,9,9,
\]
whose median is $9$, which appears in the list. If there are three remaining terms, they must be
$1,1,1$, but then $1$ and $9$ both occur three times. Thus this case would
\stweak{create another repeated mode}, so we reject it. Therefore, we
\stpos{rule out three nines case}.

Thus $9$ appears exactly twice. The two $9$'s contribute $18$, so the remaining entries sum to
$12$. Since $9$ is the unique mode, no other value can appear twice. Therefore, the remaining
entries must be distinct. We should not \stweak{allow another duplicate value}; instead, we
\stpos{enforce remaining values are distinct}.

Because the total list length is even and there are already two $9$'s, the number of remaining
entries must also be even. The remaining entries can have either two or four terms. We
\stpos{test feasible remaining length cases}.

If there are four remaining distinct positive integers summing to $12$, the only possibilities are
\[
1,2,3,6
\qquad\text{or}\qquad
1,2,4,5.
\]
Together with the two $9$'s, their medians are respectively
\[
\frac{3+6}{2}=4.5
\qquad\text{and}\qquad
\frac{4+5}{2}=4.5.
\]
These are not positive integers, so this case fails. This shows why we should not
\stweak{keep noninteger median candidates}.

Therefore, there must be exactly two remaining entries. Let them be $a<b$. Since they sum
to $12$, we have
\[
a+b=12.
\]
We now \stpos{check pairs using median condition}. The possible distinct positive pairs are
\[
(1,11),\ (2,10),\ (4,8),\ (5,7).
\]
The pairs $(1,11)$ and $(2,10)$ place the two $9$'s in the middle, giving median $9$, which
appears in the list. The pair $(4,8)$ gives median $\frac{8+9}{2}=8.5$, not an integer.
The pair $(5,7)$ gives the sorted list
\[
5,7,9,9.
\]
Its median is
\[
\frac{7+9}{2}=8,
\]
which is a positive integer and does not appear in the list. Hence we
\stpos{verify all constraints are satisfied}: the sum is $30$, the unique mode is $9$, and the
median is $8$.

Finally, we \stpos{compute required sum of squares}:
\[
5^2+7^2+9^2+9^2
=
25+49+81+81
=
236.
\]

\textbf{Answer.}
\[
\boxed{236}
\]

\end{strideexample}

\begin{strideexample}{Example 3: Strengthened and weakened 5-gram strategic patterns}
\small

\textbf{Question.}
Let $p$ be the least prime number for which there exists a positive integer $n$ such that
$n^{4}+1$ is divisible by $p^{2}$. Find the least positive integer $m$ such that $m^{4}+1$
is divisible by $p^{2}$.

\vspace{2pt}
\textbf{Response.}
We need find the smallest prime $p$ such that
\[
n^4 \equiv -1 \pmod{p^2}
\]
has a solution. A tempting but incorrect approach is to
\stweak{test primes without structural condition}. Instead, we first
\stpos{check prime two separately first}. If $p=2$, then for odd $n$,
$n^4 \equiv 1 \pmod{16}$, so $n^4+1 \equiv 2 \pmod{16}$, which is not divisible by $4$.
For even $n$, $n^4+1$ is odd. Hence $p\neq 2$.

Now let $p$ be an odd prime. If
\[
n^4 \equiv -1 \pmod p,
\]
then
\[
n^8 \equiv 1 \pmod p
\]
but $n^4 \not\equiv 1 \pmod p$. Therefore $n$ has order $8$ modulo $p$. Thus we should
\stpos{require order eight modulo prime}, rather than
\stweak{assume any odd prime works}. Since the multiplicative group modulo $p$ has order
$p-1$, we must have
\[
8 \mid p-1.
\]
Therefore, we \stpos{use cyclic group divisibility condition} and obtain
\[
p \equiv 1 \pmod 8.
\]
The least prime satisfying this is
\[
p=17.
\]

Next we need solve
\[
m^4 \equiv -1 \pmod{17^2}.
\]
We should not \stweak{only solve modulo p once}, because the problem requires divisibility by
$p^2$. First solve modulo $17$:
\[
x^4 \equiv -1 \equiv 16 \pmod{17}.
\]
The solutions modulo $17$ are
\[
x \equiv 2,8,9,15 \pmod{17}.
\]
At this point, it is wrong to \stweak{choose smallest root modulo prime}, since the least lifted
solution modulo $17^2$ may come from a different root.

We now \stpos{lift roots using Hensel condition}. Write
\[
m=a+17k,
\]
where $a$ is one of the roots modulo $17$. Let
\[
f(x)=x^4+1.
\]
Then the lifting condition is
\[
f(a)+17k f'(a)\equiv 0 \pmod{17^2}.
\]
Equivalently,
\[
\frac{f(a)}{17}+k f'(a)\equiv 0 \pmod{17}.
\]
Since
\[
f'(a)=4a^3,
\]
and none of the roots $a=2,8,9,15$ is divisible by $17$, each root lifts uniquely.

Now we \stpos{compute each lifted residue class}. For the four roots modulo $17$, the lifted
solutions modulo $289$ are
\[
2+17\cdot 9=155,
\]
\[
8+17\cdot 6=110,
\]
\[
9+17\cdot 10=179,
\]
\[
15+17\cdot 7=134.
\]
We must \stpos{compare all lifted positive residues}, rather than
\stweak{forget compare all lifted residues}. The least positive lifted solution is
\[
m=110.
\]

Finally, we \stpos{verify divisibility modulo p squared}:
\[
110^4+1 \equiv 0 \pmod{289}.
\]
Therefore, the least positive integer $m$ is
\[
110.
\]

\textbf{Answer.}
\[
\boxed{110}
\]

\end{strideexample}

\clearpage
\twocolumn

\end{document}